\documentclass[english,10pt,twocolumn,letterpaper]{article}
\usepackage[T1]{fontenc}
\usepackage[latin9]{inputenc}
\usepackage{geometry}
\usepackage{array}
\usepackage{multirow}
\usepackage{amsmath}
\usepackage{amssymb}
\usepackage{graphicx}

\makeatletter

\providecommand{\tabularnewline}{\\}
 \usepackage{cvpr}
 \usepackage{times}
 \usepackage{amsmath}
 \usepackage{amssymb}
\def\cvprPaperID{1616}
\cvprfinalcopy
\usepackage{etoolbox} 
\patchcmd\thebibliography  {\labelsep}  {\labelsep\itemsep=0pt\relax}  {}  {\typeout{Couldn't patch the command}}

\makeatother

\usepackage{babel}
\begin{document}

\title{Combining Local Appearance and Holistic View: Dual-Source Deep Neural
Networks for Human Pose Estimation}

\author{%
\begin{tabular}{>{\centering}p{3cm}||>{\centering}p{3cm}||>{\centering}p{3cm}||>{\centering}p{3cm}}
\multicolumn{4}{c}{Xiaochuan Fan, $\ $Kang Zheng, $\ $Yuewei Lin, $\ $Song Wang}\tabularnewline
\multicolumn{4}{>{\centering}p{14cm}}{{\small{}Department of Computer Science \& Engineering, University
of South Carolina, Columbia, SC 29208, USA}}\tabularnewline
\multicolumn{4}{c}{\tt{\{\small{}fan23,zheng37,lin59\}@email.sc.edu, songwang@cec.sc.edu}}\tabularnewline
\end{tabular}}

\maketitle
\thispagestyle{empty}
\begin{abstract}
We propose a new learning-based method for estimating 2D human pose
from a single image, using Dual-Source Deep Convolutional Neural Networks
(DS-CNN). Recently, many methods have been developed to estimate human
pose by using pose priors that are estimated from physiologically
inspired graphical models or learned from a holistic perspective.
In this paper, we propose to integrate both the local (body) part
appearance and the holistic view of each local part for more accurate
human pose estimation. Specifically, the proposed DS-CNN takes a set
of image patches (category-independent object proposals for training
and multi-scale sliding windows for testing)  as the input and then
learns the appearance of each local part by considering their holistic
views in the full body. Using DS-CNN, we achieve both joint detection,
which determines whether an image patch contains a body joint, and
joint localization, which finds the exact location of the joint in
the image patch. Finally, we develop an algorithm to combine these
joint detection/localization results from all the image patches for
estimating the human pose. The experimental results show the effectiveness
of the proposed method  by comparing to the state-of-the-art human-pose
estimation methods based on pose priors that are estimated from physiologically
inspired graphical models or learned from a holistic perspective.
\end{abstract}

\section{Introduction}

By accurately locating the important body joints from 2D images, human
pose estimation plays an essential role in computer vision. It has
wide applications in intelligent surveillance, video-based action
recognition, and human-computer interaction. However, human pose estimation
from an 2D image is a well known challenging problem -- too many degrees
of freedom are introduced by the large variability of the human pose,
different visual appearance of the human body and joints, different
angles of camera view, and possible occlusions of body parts and joints.

Most of the previous works on human pose estimation are based on the
two-layer part-based model \cite{Felzenszwalb2005,Tian2012,Wang2013,Eichner2013,Duan2012,Sun2011,Pishchulin2013,Andriluka2009,Pishchulin2012,Johnson2010,Sapp2010,Dantone2013,Singh2010,Wang2008,Sapp2013}.
The first layer focuses on local (body) part appearance and the second
layer imposes the contextual relations between local parts. One popular
part-based approach is pictorial structures \cite{Felzenszwalb2005},
which capture the pairwise geometric relations between adjacent parts
using a tree model. However, these pose estimation methods using part-based
models are usually sensitive to noise and the graphical model lacks
expressiveness to model complex human poses \cite{Dantone2013}. Furthermore,
most of these methods  search for each local part independently and
the local appearance may not be sufficiently discriminative for identifying
each local part reliably. 

\begin{figure}
\begin{centering}
\includegraphics[scale=0.2]{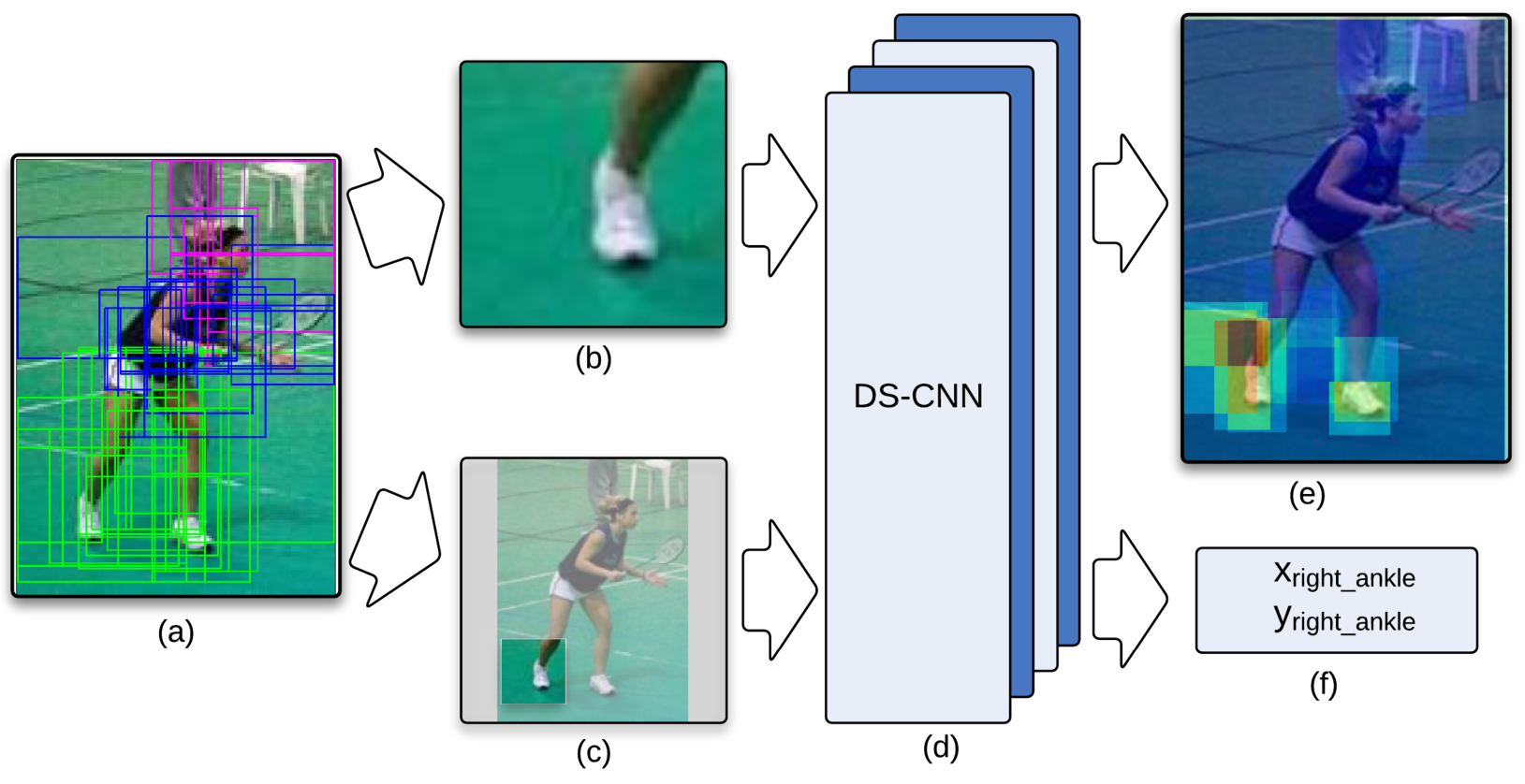}
\par\end{centering}

\caption{An illustration of the proposed method based on DS-CNN. (a) Input
image and generated image patches. (b) DS-CNN input on an image patch
(containing a local part -- ankle). (c) DS-CNN input on full body
and holistic view of the local part in the full body. (d) DS-CNN for
learning. (e) DS-CNN output on joint detection. (f) DS-CNN output
on joint localization.\label{fig:overview}}

\vspace{-0.0in}
\end{figure}

Recently, deep neural network architectures, specifically deep convolutional
neural networks (CNNs), have shown outstanding performance in many
computer vision tasks. Due to CNNs' large learning capacity and robustness
to variations, there is a natural rise in the interest to directly
learn high-level representations of human poses without using hand-crafted
low-level features and even graphical models. Toshev et al. \cite{Toshev2014}
present such a holistic-styled pose estimation method named DeepPose
using DNN-based joint regressors. This method also uses a two-layer
architecture: The first layer resolves ambiguity between body parts
(e.g. left and right legs) in a holistic way and provides an initial
pose estimation and the second layer refines the joint locations in
a local neighborhood around the initial estimation. From the experiments
in \cite{Toshev2014}, DeepPose can achieve better performance on
two widely used datasets, FLIC and LSP, than several recently developed
human pose estimation methods. However, DeepPose does not consider
local part appearance in initial pose estimation. As a result, it
has difficulty in estimating complex human poses, even using the CNN
architecture.

In this paper, we propose a dual-source CNN (DS-CNN) based method
for human pose estimation, as illustrated in Fig. \ref{fig:overview}.
This proposed method integrates both the local part appearance in
image patches and the holistic view of each local part for more accurate
human pose estimation. Following the region-CNN (R-CNN) that was developed
for object detection \cite{Girshick2014}, the proposed DS-CNN takes
a set of category-independent object proposals detected from the input
image for training. Compared to the sliding windows or the full image,
that are used as the input in many previous human pose estimation
methods, object proposals can capture the local body parts with better
semantic meanings in multiple scales \cite{Girshick2014,Zhang2014}.
In this paper, we extend the original single-source R-CNN to a dual-source
model (DS-CNN) by including the full body and the holistic view of
the local parts as a separate input, which provides a holistic view
for human pose estimation. By taking both the local part object proposals
and the full body as inputs in the training stage, the proposed DS-CNN
performs a unified learning to achieve both joint detection, which
determines whether an object proposal contains a body joint, and joint
localization, which finds the exact location of the joint in the object
proposal. In the testing stage, we use multi-scale sliding windows
to provide local part information in order to avoid the performance
degradation resulted by the uneven distribution of object proposals.
Based on the DS-CNN outputs, we combine the joint detection results
from all the sliding windows to construct a heatmap that reflects
the joint location likelihood at each pixel and weightedly average
the joint localization results at the high-likelihood regions of the
heatmap to achieve the final estimation of each joint location.

In the experiments, we test the proposed method on two widely used
datasets and compare its performance to several recently reported
human pose estimation methods, including DeepPose. The results show
the effectiveness of the proposed method which combines local appearance
and holistic view.

\section{Related Work}

\textbf{Part-based models for human pose estimation.} In the part-based
models, human body is represented by a collection of physiologically
inspired parts assembled through a deformable configuration. Following
the pictorial-structure model \cite{Fischler1973,Felzenszwalb2005},
a variety of part-based methods have been developed for human pose
estimation \cite{Tian2012,Wang2013,Eichner2013,Duan2012,Sun2011,Pishchulin2013,Andriluka2009,Pishchulin2012,Johnson2010,Sapp2010,Dantone2013,Singh2010,Wang2008,Sapp2013}.
While many early methods build appearance models for each local part
independently, recent works \cite{Andriluka2009,Eichner2009,Eichner2013,Dantone2013,Johnson2010,Sapp2013}
attempt to design strong body part detectors by capturing the contextual
relations between body parts. Johnson and Everingham \cite{Johnson2010}
partition the pose space into a set of pose clusters and then apply
nonlinear classifiers to learn pose-specific part appearance. In \cite{Dantone2013},
independent regressors are trained for each joint and the results
from these regressors are combined to estimate the likelihood of each
joint at each pixel of the image. Based on the appearance models built
for each part, these methods usually leverage tree-structured graphical
models to further impose the pairwise geometric constraints between
parts \cite{Tian2012,Wang2013,Andriluka2009,Pishchulin2013,Yang2011}.
Due to the limited expressiveness \cite{Toshev2014}, the tree-structured
graphical models often suffer from the limb ambiguity, which affects
the accuracy of human pose estimation. There have been several works
that focus on designing richer graphical models to overcome the limitation
of tree-structured graphical models. For example, in \cite{Johnson2010},
mixture of pictorial structure models are learned to capture the \textquoteleft{}multi-modal\textquoteright{}
appearance of each body part. Yang and Ramanan \cite{Yang2011} introduce
a flexible mixture-of-parts model to capture contextual co-occurrence
relations between parts. In \cite{Tian2012}, the hierarchical structure
is incorporated to model high-order spatial relation among parts.
Loopy models \cite{Duan2012,Jiang2008,Tian2010,Ren2005} allow to
include additional part constraints, but require approximate inference.
In the latter experiments, we include several above-mentioned part-based
methods for performance comparison.

\textbf{Deep convolutional neural network (CNN)} \textbf{in computer
vision.} As a popular deep learning approach, CNN \cite{LeCun1990}
attempts to learn multiple levels of representation and abstraction
and then use it to model complex non-linear relations. It has been
shown to be a useful tool in many computer vision applications. For
example, it has demonstrated impressive performance for image classification
\cite{Jarrett2009,LeCun2004,Lee2009,Krizhevsky2012}. More recently,
CNN architectures have been successfully applied to object localization
and detection \cite{Szegedy2013,Girshick2014,Sermanet2014}. In \cite{Sermanet2014},
a single shared CNN named `Overfeat' is used to simultaneously classify,
locate and detect objects from an image by examining every sliding
window. In this paper, we also integrate joint detection and localization
using a single DS-CNN. But our problem is much more challenging than
object detection -- we need to find precise locations of a set of
joints for human pose estimation. Girshick et al. \cite{Girshick2014}
apply high-capacity R-CNNs to bottom-up object proposals \cite{Uijlings2013}
for object localization and segmentation. It achieves 30\% performance
improvement on PASCAL VOC 2012 against the state of the art. Zhang
et al. \cite{Zhang2014} adopt the R-CNN \cite{Girshick2014} to part
localization and verify that the use of object proposals instead of
sliding windows in CNN can help localize smaller parts. Based on this,
R-CNN is shown to be effective for fine-grained category detection.
However, this method does not consider the complex relations between
different parts \cite{Zhang2014} and is not applicable to human pose
estimation.

\textbf{CNN for human pose estimation.} In \cite{Toshev2014}, a cascade
of CNN-based joint regressors are applied to reason about pose in
a holistic manner and the developed method was named `DeepPose'. The
DeepPose networks take the full image as the input and output the
ultimate human pose without using any explicit graphical model or
part detectors.

In \cite{Jain2014}, Jain et al. introduce a CNN-based architecture
and a learning technique that learns low-level features and a higher-level
weak spatial model. Following \cite{Jain2014}, Tompson et al. show
that the inclusion of a MRF-based graphical model into the CNN-based
part detector can substantially increase the human pose estimation
performance. Different from DeepPose and Tompson et al. \cite{Tompson2014},
the proposed method takes both the object proposals and the full body
as the input for training, instead of using the sliding-windowed patches,
to capture the local body parts with better semantic meanings in multiple
scales.

\section{Problem Description and Notations}

In this paper, we adopt the following notations. A human pose can
be represented by a set of human joints $\mathbf{J}=\left\{ \mathbf{j}_{i}\right\} _{i=1}^{L}\in\mathbb{R}^{2L\times1}$,
where $\mathbf{j}_{i}=\left(x_{i},y_{i}\right)^{T}$ denotes the 2D
coordinate of the joint $i$ and $L$ is the number of human joints.
In this paper, we are interested in estimating the 2D joint locations
$\mathbf{J}$ from a single image $I$. Since our detection and regression
are applied to a set of image patches, in the form of rectangular
bounding boxes, detected in $I$, it is necessary to convert absolute
joint coordinates in image $I$ to relative joint coordinates in an
image patch. Furthermore, we introduce a normalization to make the
locations invariant to size of different image patches. Specifically,
given an image patch $\mathbf{p}$, the location of $\mathbf{p}$
is represented by 4-element vector $\mathbf{p}=\left(w\left(\mathbf{p}\right),h\left(\mathbf{p}\right),\mathbf{c}\left(\mathbf{p}\right)\right)^{T}$,
where $w\left(\mathbf{p}\right)$ and $h\left(\mathbf{p}\right)$
are the width and height of $\mathbf{p}$, $\mathbf{c}\left(\mathbf{p}\right)=\left(x_{c}\left(\mathbf{p}\right),y_{c}\left(\mathbf{p}\right)\right)^{T}$
is the center of $\mathbf{p}$. Then the normalized coordinate of
joint $\mathbf{j}_{i}$ relative to $\mathbf{p}$ can be denoted as
\begin{align}
\mathbf{j}_{i}\left(\mathbf{p}\right) & =\left(x_{i}\left(\mathbf{p}\right),y_{i}\left(\mathbf{p}\right)\right)^{T}\nonumber \\
 & =\left(\frac{x_{i}-x_{c}\left(\mathbf{p}\right)}{w\left(\mathbf{p}\right)},\frac{y_{i}-y_{c}\left(\mathbf{p}\right)}{h\left(\mathbf{p}\right)}\right)^{T}.\label{eq:cal_relative_coordinates}
\end{align}

Furthermore, the visibility of all the joints in $\mathbf{p}$ is
denoted as $\mathbf{V}\mbox{\ensuremath{\left(\mathbf{p}\right)}}=\left\{ v_{i}\left(\mathbf{p}\right)\right\} _{i=1}^{L}\in\mathbb{R}^{L\times1}$,
where 
\begin{equation}
v_{i}\left(\mathbf{p}\right)=\begin{cases}
1, & \left|x_{i}\left(\mathbf{p}\right)\right|\leq0.5\text{ and}\left|y_{i}\left(\mathbf{p}\right)\right|\leq0.5\\
0, & \text{otherwise}.
\end{cases}\label{eq:joint_vis}
\end{equation}

If $v_{i}\left(\mathbf{p}\right)=1$, it indicates that the joint
$i$ is visible in $\mathbf{p}$, i.e., it is located inside the patch
$\mathbf{p}$. On the contrary, if $v_{i}\left(\mathbf{p}\right)=0$,
it indicates that the joint $i$ is invisible in $\mathbf{p}$, i.e.,
it is located outside of $\mathbf{p}$.

\begin{figure}
\begin{centering}
\includegraphics[scale=0.12]{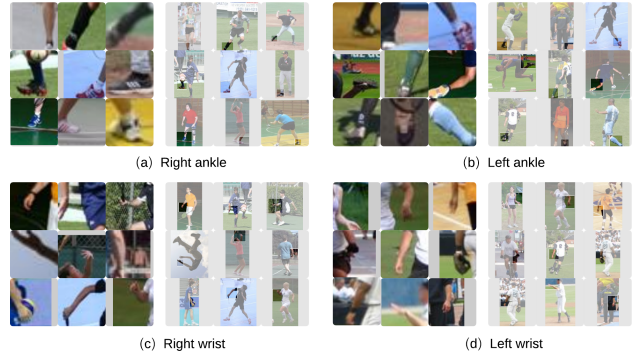}
\par\end{centering}

\caption{Extended part and body patches containing (a) right ankle, (b) left
ankle, (c) right wrist, and (d) left wrist from the LSP training dataset.
For each local part, the part patches are shown in the left while
the corresponding body patches are shown in the right. \label{fig:Extended-part-and-global-proposal}}

\vspace{-0.1in}
\end{figure}

\section{Model Inputs\label{sec:Model-Inputs}}

As described earlier, to combine the local part appearance and the
holistic view of each part, the proposed DS-CNN takes two inputs for
training and testing -- image patches and the full body. To make it
clearer, we call the former input the \textit{part patches,} denoted
as $\mathbf{p}_{p}$, and the latter \textit{body patches}, denoted
as $\mathbf{p}_{b}$. So the dual-source input is $\mathbf{p}_{p,b}=\left(\mathbf{p}_{p},\mathbf{p}_{b}\right)$.
Randomly selected samples for these two kinds of inputs are shown
in Fig. \ref{fig:Extended-part-and-global-proposal}, where for each
local part, the part patches are shown in the left while the corresponding
body patches are shown in the right. From these samples, we can see
that it is difficult to distinguish the left and right wrists, or
some wrists and legs, based only on the local appearance in the part
patches.

As we will see later, we use object proposals detected from an image
for training and object proposals usually show different sizes and
different aspect ratios. CNN requires the input to be of a fixed dimension.
In \cite{Girshick2014}, all the object proposals are non-uniformly
scaled to a fixed-size square and it may need to vary the original
aspect ratios. This may complicate the CNN training by artificially
introducing unrealistic patterns into training samples. In particular,
in our model we are only interested in the body joint that is closest
to the center of a part patch (This will be elaborated in detail later).
If the part patch is non-uniformly scaled, the joint of interest may
be different after the change of the aspect ratio. Thus, in this paper
we keep the aspect ratio of image patches unchanged when unifying
its size. Specifically, we extend the short side of the image patch
to include additional rows or columns to make it a square. This extension
is conducted in a way such that the center of each image patch keeps
unchanged. After the extension, we can perform uniform scaling to
make each patch a fixed-size square. This extension may not be preferred
in object detection \cite{Girshick2014} by including undesired background
information. However, in our problem this extension just includes
more context information of the joint of interest. This will not introduce
much negative effect to the part detection. The only minor effect
is the a subtle reduction of the  resolution of each patch (after
the uniform scaling).

\textbf{Part Patches }In the training stage, we construct part patches
in two steps. 1) Running an algorithm to construct a set of category-independent
object proposals. Any existing object proposal algorithms can be used
here. In our experiments, we use the algorithm developed in \cite{Zitnick2014}.
2) Select a subset of the constructed proposals as the part patches.
We consider two factors for Step 2). First, we only select object
proposals with a size in certain range as part patches. If the size
of an object proposal is too large, it may cover multiple body parts
and its appearance lacks sufficient resolution (after the above-mentioned
uniform scaling) for joint detection and localization. On the contrary,
if the size of an object proposal is too small, its appearance may
not provide sufficient features. To address this issue, we only select
the object proposals $\mathbf{p}_{p}$ with an area in a specified
range as part patches, i.e., 
\begin{equation}
\mu_{1}d^{2}\left(\mathbf{J}\right)\leq w\left(\mathbf{p}_{p}\right)\cdot h\left(\mathbf{p}_{p}\right)\leq\mu_{2}d^{2}\left(\mathbf{J}\right)\label{eq:part_proposal_upper_bound}
\end{equation}
where $d\left(\mathbf{J}\right)$ is the distance between two opposing
joints on the human torso \cite{Toshev2014}. $\mu_{1}$ and $\mu_{2}$
are two coefficients ($\mu_{1}$<$\mu_{2}$) that help define the
lower bound and the upper bound for selecting an object proposal as
a part patch.

\begin{figure}
\begin{centering}
\includegraphics[scale=0.19]{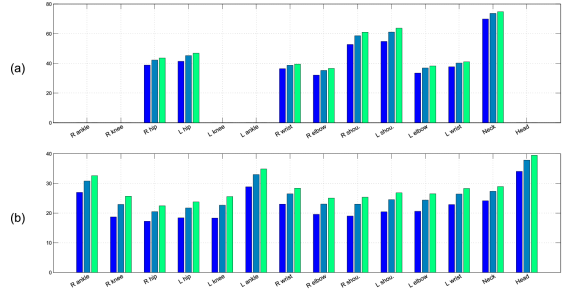}
\par\end{centering}

\caption{The average number of part patches that cover each joint in (a) FLIC
and (b) LSP datasets. Three colors indicates the results by selecting
different $\mu_{2}$ values of 1.0, 1.5 and 2.0 respectively. \label{fig:Joint-histogram}}

\vspace{-0.1in}
\end{figure}

Second, from the training perspective, we desire all the body joints
are covered by sufficient number of part patches. In the ideal case,
we expect the selected part patches cover all the joints in a balanced
way -- all the joints are covered by similar number of part patches.
We empirically examine this issue and results are shown in Fig. \ref{fig:Joint-histogram}
-- on both FLIC and LSP datasets, this simple part-patch selection
algorithm provides quite balanced coverage to all the joints. In this
figure, the x-axis indicates the label of different joints -- only
upper-body joints are shown in FLIC dataset while all 14 body joints
are shown in LSP dataset. The y-axis indicates the average number
of part patches that covers the specified joint in each image. Here
we count that a part patch covers a joint if this joint is visible
to (i.e., located inside) this patch and this joint is the closest
joint to the center of this patch. At each joint, we show three part-patch
coverage numbers in three different colors. From left to right, they
correspond to three different $\mu_{2}$ values of 1.0, 1.5 and 2.0
respectively. In this empirically study, we always set $\mu_{1}=0.1$. 

In the testing stage, part patches are selected from multi-scale sliding
windows (this will be justified in Section \ref{sec:Experiments}).

\begin{figure*}
\begin{centering}
\includegraphics[scale=0.56]{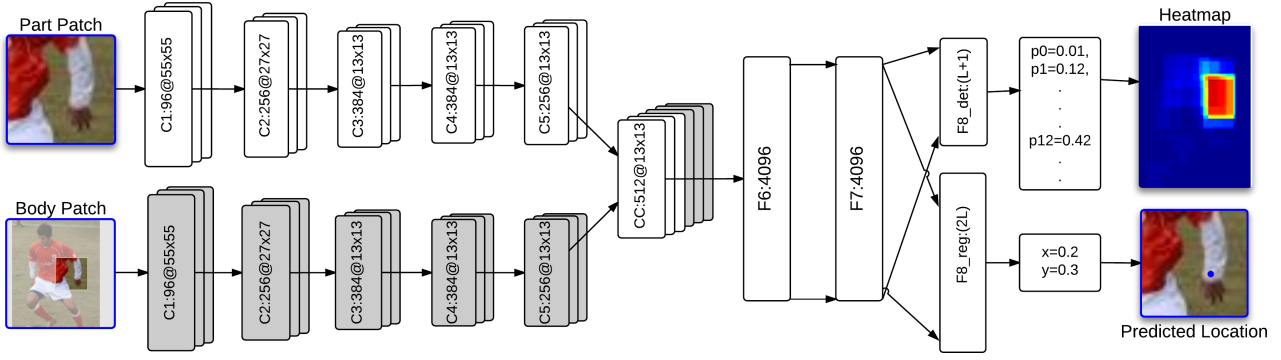}
\par\end{centering}

\caption{The structure of DS-CNN. \label{fig:network_architecture}}

\vspace{-0.1in}
\end{figure*}

\textbf{Body Patches} Similarly, in the training stage we construct
body patches by selecting a subset of object proposals from the same
pool of object proposals detected from the image. The only requirement
is that the selected body patch should cover the whole body or all
the joints, i.e., 
\begin{equation}
\sum_{i=1}^{L}v_{i}\left(\mathbf{p}_{b}\right)=L.\label{eq:body_proposal_condition}
\end{equation}

In the testing stage, the body patch can be generated by using a human
detector. For the experiments in this paper, each testing image only
contains one person and we simply take the whole testing image as
the body patch.

For DS-CNN, each training sample is made up of a part patch $\mathbf{p}_{p}$,
a body patch $\mathbf{p}_{b}$, and the binary mask that specifies
the location of $\mathbf{p}_{p}$ in $\mathbf{p}_{b}$, as shown in
Fig. \ref{fig:Extended-part-and-global-proposal}, where both $\mathbf{p}_{p}$
and $\mathbf{p}_{b}$ are extended and normalized to a fixed-size
square image. For part patch $\mathbf{p}_{p}$, we directly take the
RGB values at all the pixels as the first source of input to DS-CNN.
For body patch $\mathbf{p}_{b}$, we take the binary mask as an additional
alpha channel and concatenate the RGB values of $\mathbf{p}_{b}$
and the alpha values as the second source of input to DS-CNN. Given
that we extend and normalize all the patches to a $N\times N$ square,
the first source of input is of dimension $3N^{2}$ and the second
source of input is of size $4N^{2}.$ In the training stage, based
on the constructed part patches and body patches, we randomly select
one from each as a training sample. For both training and testing,
if the selected part patch is not fully contained in the selected
body patch, we crop the part patch by removing the portion located
outside the body patch before constructing the training or testing
sample.

\section{Multi-Task Learning\label{sec:Multi-Task-Learning}}

We combine two tasks in a single DS-CNN -- joint detection, which
determines whether a part patch contains a body joint, and joint localization,
which finds the exact location of the joint in the part patch. Each
task is associated with a loss function.

\textbf{Joint detection} For joint detection, we label a patch-pair
$\mathbf{p}_{p,b}$ to joint $i^{*}$, where 

\begin{equation}
i^{*}\left(\mathbf{p}_{p}\right)=\begin{cases}
\text{arg }\underset{1\leq i\leq L}{\text{min}}\parallel\mathbf{j}_{i}\left(\mathbf{p}_{p}\right)\parallel^{2} & \text{if }\sum_{k=1}^{L}v_{k}\left(\mathbf{p}_{p}\right)>0\\
0 & \text{otherwise},
\end{cases}\label{eq:part_proposal_label}
\end{equation}
and this is taken as the ground truth for training.

Let the DS-CNN output for joint detection be $\left(\mathcal{\ell}_{0}\left(\mathbf{p}_{p,b}\right),\mathcal{\ell}_{1}\left(\mathbf{p}_{p,b}\right)...,\mathcal{\ell}_{L}\left(\mathbf{p}_{p,b}\right)\right)^{T}$,
where $\mathcal{\ell}_{0}$ indicates the likelihood of no body joint
visible in $\mathbf{p}_{p}$ and $\mathcal{\ell}_{i},i=1,...,L$ represents
the likelihood that joint $i$ is visible in $\mathbf{p}_{p}$ and
is the closest joint to the center of $\mathbf{p}_{p}$ . We use a
softmax classifier where the loss function is 
\begin{equation}
C_{d}\left(\mathbf{p}_{p,b}\right)=-\sum_{i=0}^{L}1\left(i^{*}\left(\mathbf{p}_{p}\right)=i\right)\text{log}\left(\ell_{i}\left(\mathbf{p}_{p,b}\right)\right),\label{eq:part_proposal_joint_location}
\end{equation}
 where $1\left(\cdot\right)$ is the indicator function.

\textbf{Joint localization} Joint localization is formulated as a
regression problem. In DS-CNN training, the ground-truth joint location
for a patch-pair $\mathbf{p}_{p,b}$ is $\mathbf{j}_{i^{*}\left(\mathbf{p}_{p}\right)}\left(\mathbf{p}_{p}\right)=\left(x_{i^{*}\left(\mathbf{p}_{p}\right)}\left(\mathbf{p}_{p}\right),y_{i^{*}\left(\mathbf{p}_{p}\right)}\left(\mathbf{p}_{p}\right)\right)^{T}$,
where $i^{*}\left(\mathbf{p}_{p}\right)$ is defined in Eq. (\ref{eq:part_proposal_label}).
Let the DS-CNN output on joint localization be $\left\{ \mathbf{z}_{i}\left(\mathbf{p}_{p,b}\right)\right\} _{i=1}^{L}\in\mathbb{R}^{2L\times1}$
, where $\mathbf{z}_{i}\left(\mathbf{p}_{p,b}\right)=\left(\widehat{x}_{i},\widehat{y}_{i}\right)^{T}$
denotes the predicted location of the $i-$th joint in $\mathbf{p}_{p}$.
We use the mean squared error as the loss function, 
\begin{equation}
C_{r}\left(\mathbf{p}_{p,b}\right)=\begin{cases}
\parallel\mathbf{z}_{i^{*}\left(\mathbf{p}_{p}\right)}\left(\mathbf{p}_{p,b}\right)-\mathbf{j}_{i^{*}\left(\mathbf{p}_{p}\right)}\left(\mathbf{p}_{p}\right)\parallel^{2} & \text{if }i^{*}>0\\
0 & \text{otherwise}.
\end{cases}\label{eq:regession_cost}
\end{equation}

Combining the joint detection and joint localization, the loss function
for DS-CNN is 
\begin{align}
C & =\sum_{\mathbf{p}_{p,b}}\left\{ \lambda_{d}C_{d}\left(\mathbf{p}_{p,b}\right)+C_{r}\left(\mathbf{p}_{p,b}\right)\right\} ,\label{eq:total_cost}
\end{align}
where the summation is over all the training samples (patch pairs)
and $\lambda_{d}>0$ is a factor that balances the two loss functions.

\section{DS-CNN Structure}

The structure of the proposed DS-CNN is based on the CNN described
in \cite{Krizhevsky2012}, which is made up of five convolutional
layers, three fully-connected layers, and a final 1000-way softmax,
in sequence. The convolutional layers 1, 2 and 5 are followed by max
pooling. In the proposed DS-CNN, we include two separate sequence
of convolutional layers. As shown in Fig. \ref{fig:network_architecture},
one sequence of 5 convolutional layers takes the part-patch input
as defined in Section \ref{sec:Model-Inputs} and extracts the features
from local appearance. The other sequence of 5 convolutional layers
takes the body-patch input and extracts the holistic features of each
part. The output from these two sequences of convolutional layers
are then concatenated together, which are then fed to a sequence of
three fully connected layers. We replace the final 1000-way softmax
by a $\left(L+1\right)$-way softmax and a regressor for joint detection
and joint localization, respectively. In DS-CNN, all the convolutional
layers and the fully-connected layers are shared by both the joint
detection and the joint localization. 

In Fig. \ref{fig:network_architecture}, the convolutional layer and
the following pooling layer is labeled ${\tt C_{i}}$ and the fully-connected
layer are labeled as  ${\tt F_{i}}$ where $i$ is the index of layer.
The size of a convolutional layer is described as $\text{depth}@\text{width}\times\text{height}$,
where depth is the number of convolutional filters, width and height
denote the spatial dimension.

\section{Human Pose Estimation \label{sec:Human-Pose-Estimation}}

Given a testing image, we construct a set of patch-pairs using multi-scale
sliding windows as discussed in Section \ref{sec:Model-Inputs}. We
then run the trained DS-CNN on each patch-pair $\mathbf{p}_{p,b}$
to obtain both joint detection and localization. In this section,
we propose an algorithm for estimating the final human pose on the
testing image by combining the joint detection and localization results
from all the patch pairs. 

At first, we construct a heatmap $H_{i}$ for each joint $i$ -- the
heatmap is of the same size as the original image and $H_{i}(\mathbf{x})$,
the heatmap value at a pixel $\mathbf{x},$ reflects the likelihood
that joint $i$ is located at $\mathbf{\mathbf{x}}$. Specifically,
for each patch-pair $\mathbf{p}_{p,b}$, we uniformly allocate its
joint-detection likelihood to all the pixels in $\mathbf{p}_{p}$,
i.e.,

\begin{equation}
h_{i}\left(\mathbf{x},\mathbf{p}_{p,b}\right)=\begin{cases}
\mathcal{\mathcal{\ell}}_{i}\left(\mathbf{p}_{p,b}\right)/\left(w\left(\mathbf{p}_{p}\right)\cdot h\left(\mathbf{p}_{p}\right)\right),\\
\mathbf{\text{if }x}\in\mathbf{p}_{p}\text{\ and\ }\ell_{i}\left(\mathbf{p}_{p,b}\right)>\ell_{j}\left(\mathbf{p}_{p,b}\right),\forall j\neq i\\
0\ \ \ \ \ \ \ \ \ \ \ \ \ \ \ \ \ \ \ \ \text{otherwise}. & \text{}
\end{cases}\label{eq:vote_of_a_proposal}
\end{equation}
We then sum up the allocated joint-detection likelihood over all the
patch-pairs in a testing image as 
\begin{equation}
H_{i}\left(\mathbf{x}\right)=\sum_{\mathbf{p}_{p,b}}h_{i}\left(\mathbf{x},\mathbf{p}_{p,b}\right).\label{eq:accumulated_votes}
\end{equation}
Figure \ref{fig:network_architecture} shows an example of the heatmap
for the left wrist. We can see that, by incorporating the body patches,
the constructed heatmap resolves the limb ambiguity. However, while
the heatmap provides a rough estimation of the joint location, it
is insufficient to accurately localize the body joints.

To find the accurate location of a joint, we take the DS-CNN joint-localization
outputs from a selected subset of patch-pairs, where the joint is
visible with high likelihood. We then take the weighted average of
these selected outputs as the final location of the joint. More specifically,
we only select patch pairs $\mathbf{p}_{p,b}$ that satisfy the following
conditions when finding the location of joint $i$ in the testing
image.
\begin{enumerate}
\item The likelihood that no body joint is visible in $\mathbf{p}_{p}$
is smaller than the likelihood that joint $i$ is visible in the part
patch, i.e. 
\begin{equation}
\ell_{0}\left(\mathbf{p}_{p,b}\right)<\ell_{i}\left(\mathbf{p}_{p,b}\right).\label{eq:rule1}
\end{equation}

\item The likelihood that joint $i$ is visible in $\mathbf{p}_{p}$ should
be among the $k$ largest ones over all $L$ joints. In a special
case, if we set $k$=1, this condition requires $\ell_{i}\left(\mathbf{p}_{p,b}\right)>\ell_{j}\left(\mathbf{p}_{p,b}\right),\forall j\neq i$.
\item The maximum heatmap value (for joint $i$) in $\mathbf{p}_{p}$ is
close to the maximum heatmap value over the body patch (full testing
image in our experiments). Specifically, let 
\begin{equation}
H_{i}^{p}=\underset{\mathbf{x}\in\mathbf{p}_{p}}{\text{max }}H_{i}\left(\mathbf{x}\right),\label{eq:rule2}
\end{equation}
and 
\begin{equation}
H_{i}^{b}=\underset{\mathbf{x}\in\mathbf{p}_{b}}{\text{max }}H_{i}\left(\mathbf{x}\right).\label{eq:rule3}
\end{equation}
 We require 
\begin{equation}
H_{i}^{p}>\lambda_{h}H_{i}^{b},\label{eq:rule4}
\end{equation}
where $\lambda_{h}$ is a scaling factor between $0$ and $1$. In
our experiments, we set $\lambda_{h}=0.9$. 
\end{enumerate}
Let $\mathbf{P}^{i}$ be the set of the selected patch-pairs that
satisfy these three conditions. We estimate the location of joint
$i$ by 
\begin{equation}
\mathbf{j}'_{i}=\frac{\sum_{\mathbf{p}_{p,b}\in\mathbf{P}^{i}}\mathbf{z}'_{i}\left(\mathbf{p}_{p,b}\right)\ell_{i}\left(\mathbf{p}_{p,b}\right)}{\sum_{\mathbf{p}_{p,b}\in\mathbf{P}^{i}}\ell_{i}\left(\mathbf{p}_{p,b}\right)},\label{eq:rule5}
\end{equation}
where $\mathbf{z}'_{i}\left(\mathbf{p}_{p,b}\right)$ is the DS-CNN
estimated joint-$i$ location in the coordinates of the body patch
$\mathbf{p}_{b}$. As mentioned earlier, in our experiments, each
testing image only contains one person and we simply take the whole
image as the body patch $\mathbf{p}_{b}$, so $\mathbf{z}'_{i}\left(\mathbf{p}_{p,b}\right)$
can be derived from the DS-CNN joint localization output $\mathbf{z}{}_{i}\left(\mathbf{p}_{p}\right)$
by applying the inverse transform of Eq. (\ref{eq:cal_relative_coordinates}).

\section{Experiments \label{sec:Experiments}}

In this paper, we evaluate the proposed method on Leeds Sports Pose
(LSP) dataset \cite{Johnson2010}, the extended LSP dataset \cite{Johnson2011},
and Frames Labeled in Cinema (FLIC) dataset \cite{Sapp2013}. LSP
and its extension contains 11,000 training and 1,000 testing images
of sports people gathered from Flickr with 14 full body joints annotated.
These images are challenging because of different appearances and
strong articulation. The images in LSP dataset have been scaled so
that the most prominent person is about 150 pixels high. FLIC dataset
contains 3,987 training and 1,016 testing images from Hollywood movies
with 10 upper body joints annotated. The images in FLIC dataset contain
people with diverse poses and appearances and are biased towards front-facing
poses. 

\begin{table}
\begin{centering}
{\scriptsize{}}%
\begin{tabular}{|c|>{\centering}p{0.35cm}|>{\centering}p{0.35cm}|>{\centering}p{0.35cm}|>{\centering}p{0.35cm}|>{\centering}p{0.4cm}|>{\centering}p{0.4cm}|>{\centering}p{0.4cm}|}
\hline 
{\scriptsize{}Method} & \multicolumn{2}{c|}{{\scriptsize{}Arm}} & \multicolumn{2}{c|}{{\scriptsize{}Leg}} & \multirow{2}{0.4cm}{\centering{}{\scriptsize{}Torso}} & \multirow{2}{0.4cm}{\centering{}{\scriptsize{}Head}} & \multirow{2}{0.4cm}{\centering{}{\scriptsize{}Avg.}}\tabularnewline
\cline{2-5} 
 & \centering{}{\tiny{}Upper} & \centering{}{\tiny{}Lower} & \centering{}{\tiny{}Upper} & \centering{}{\tiny{}Lower} &  &  & \tabularnewline
\hline 
\hline 
{\scriptsize{}DS-CNN} & \centering{}\textbf{\scriptsize{}0.80} & \centering{}\textbf{\scriptsize{}0.63} & \centering{}\textbf{\scriptsize{}0.90} & \centering{}\textbf{\scriptsize{}0.88} & \centering{}\textbf{\scriptsize{}0.98} & \centering{}{\scriptsize{}0.85} & \centering{}\textbf{\scriptsize{}0.84}\tabularnewline
\hline 
\hline 
{\scriptsize{}DeepPose \cite{Toshev2014}} & \centering{}{\scriptsize{}0.56} & \centering{}{\scriptsize{}0.38} & \centering{}{\scriptsize{}0.77} & \centering{}{\scriptsize{}0.71} & \centering{}{\scriptsize{}-} & \centering{}{\scriptsize{}-} & \centering{}{\scriptsize{}-}\tabularnewline
\hline 
{\scriptsize{}Dantone et al. \cite{Dantone2013}} & \centering{}{\scriptsize{}0.45} & \centering{}{\scriptsize{}0.25} & \centering{}{\scriptsize{}0.68} & \centering{}{\scriptsize{}0.61} & \centering{}{\scriptsize{}0.82} & \centering{}{\scriptsize{}0.79} & \centering{}{\scriptsize{}0.60}\tabularnewline
\hline 
{\scriptsize{}Tian et al. \cite{Tian2012}} & \centering{}{\scriptsize{}0.52} & \centering{}{\scriptsize{}0.33} & \centering{}{\scriptsize{}0.70} & \centering{}{\scriptsize{}0.60} & \centering{}{\scriptsize{}0.96} & \centering{}\textbf{\scriptsize{}0.88} & \centering{}{\scriptsize{}0.66}\tabularnewline
\hline 
{\scriptsize{}Johnson et al. \cite{Johnson2011}} & \centering{}{\scriptsize{}0.54} & \centering{}{\scriptsize{}0.38} & \centering{}{\scriptsize{}0.75} & \centering{}{\scriptsize{}0.67} & \centering{}{\scriptsize{}0.88} & \centering{}{\scriptsize{}0.75} & \centering{}{\scriptsize{}0.66}\tabularnewline
\hline 
{\scriptsize{}Wang et al. \cite{Wang2013}} & \centering{}{\scriptsize{}0.49} & \centering{}{\scriptsize{}0.32} & \centering{}{\scriptsize{}0.74} & \centering{}{\scriptsize{}0.70} & \centering{}{\scriptsize{}0.92} & \centering{}{\scriptsize{}0.86} & \centering{}{\scriptsize{}0.67}\tabularnewline
\hline 
{\scriptsize{}Pishchulin et al. \cite{Pishchulin2013}} & \centering{}{\scriptsize{}0.54} & \centering{}{\scriptsize{}0.34} & \centering{}{\scriptsize{}0.76} & \centering{}{\scriptsize{}0.68} & \centering{}{\scriptsize{}0.88} & \centering{}{\scriptsize{}0.78} & \centering{}{\scriptsize{}0.66}\tabularnewline
\hline 
{\scriptsize{}Pishchulin et al. \cite{Pishchulin2013a}} & \centering{}{\scriptsize{}0.62} & \centering{}{\scriptsize{}0.45} & \centering{}{\scriptsize{}0.79} & \centering{}{\scriptsize{}0.73} & \centering{}{\scriptsize{}0.89} & \centering{}{\scriptsize{}0.86} & \centering{}{\scriptsize{}0.72}\tabularnewline
\hline 
\end{tabular}
\par\end{centering}{\scriptsize \par}

\centering{}
\vspace{0.1in}
\caption{PCP comparison on LSP. Note that DS-CNN, DeepPose \cite{Toshev2014}
and Johnson et al. \cite{Johnson2011} are trained with both the LSP
and its extension, while the other methods use only LSP .\label{tab:PCP-comparison}}
\end{table}

\begin{figure}
\begin{centering}
\includegraphics[width=0.23\textwidth]{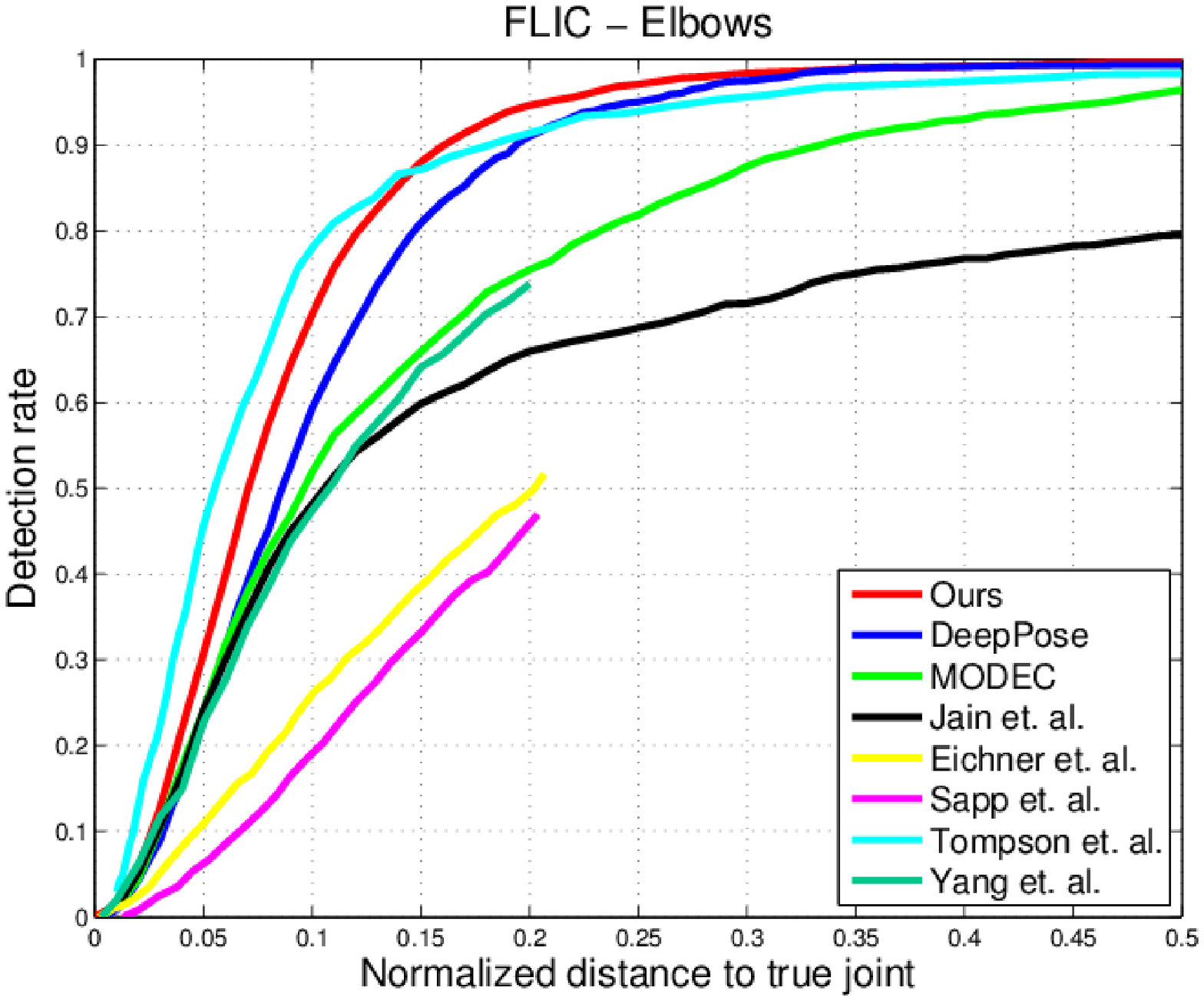}\includegraphics[width=0.23\textwidth]{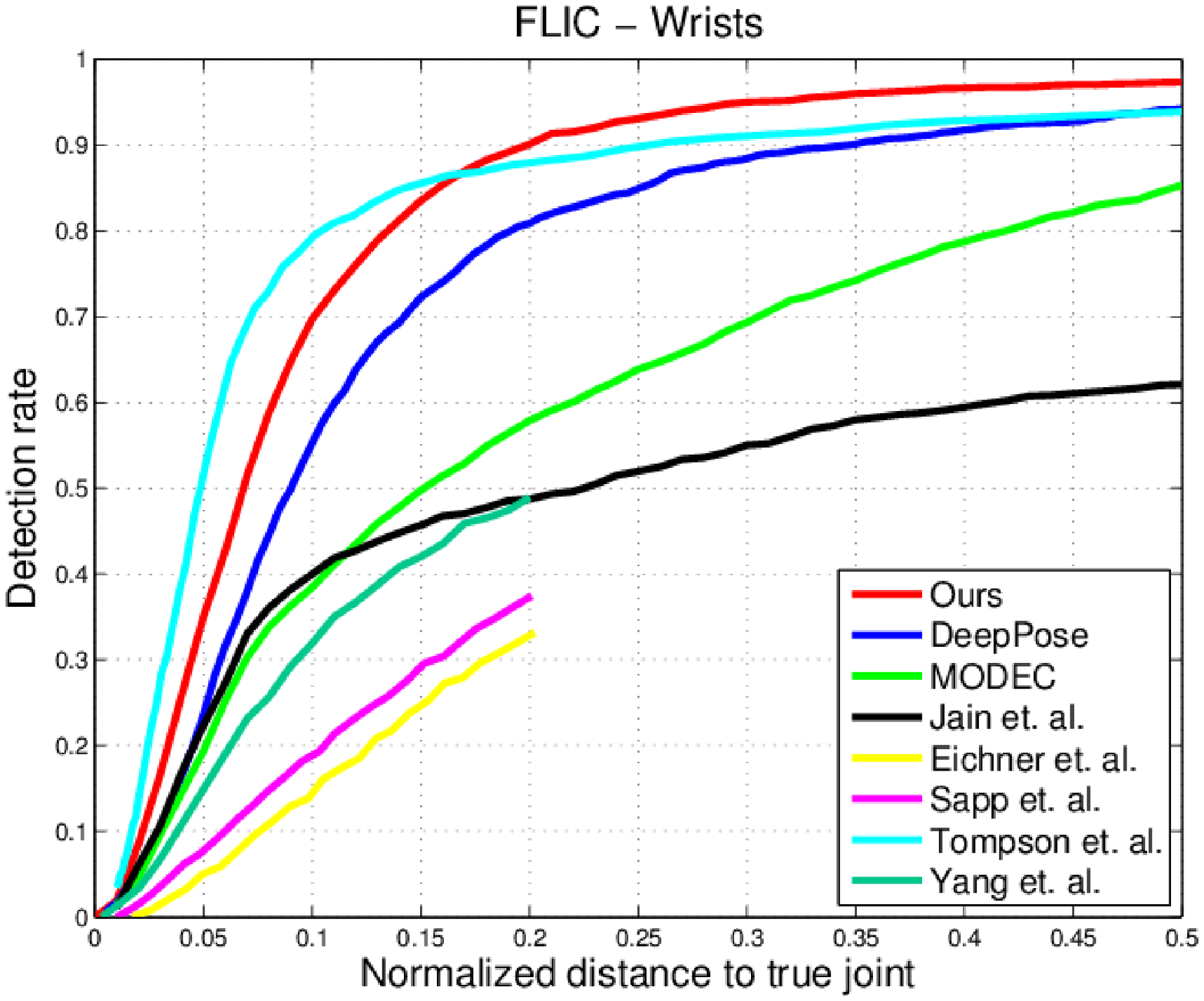}\caption{PDJ comparison on FLIC. \label{fig:PDJs_on_FLIC}}

\par\end{centering}

\begin{centering}
\includegraphics[width=0.23\textwidth]{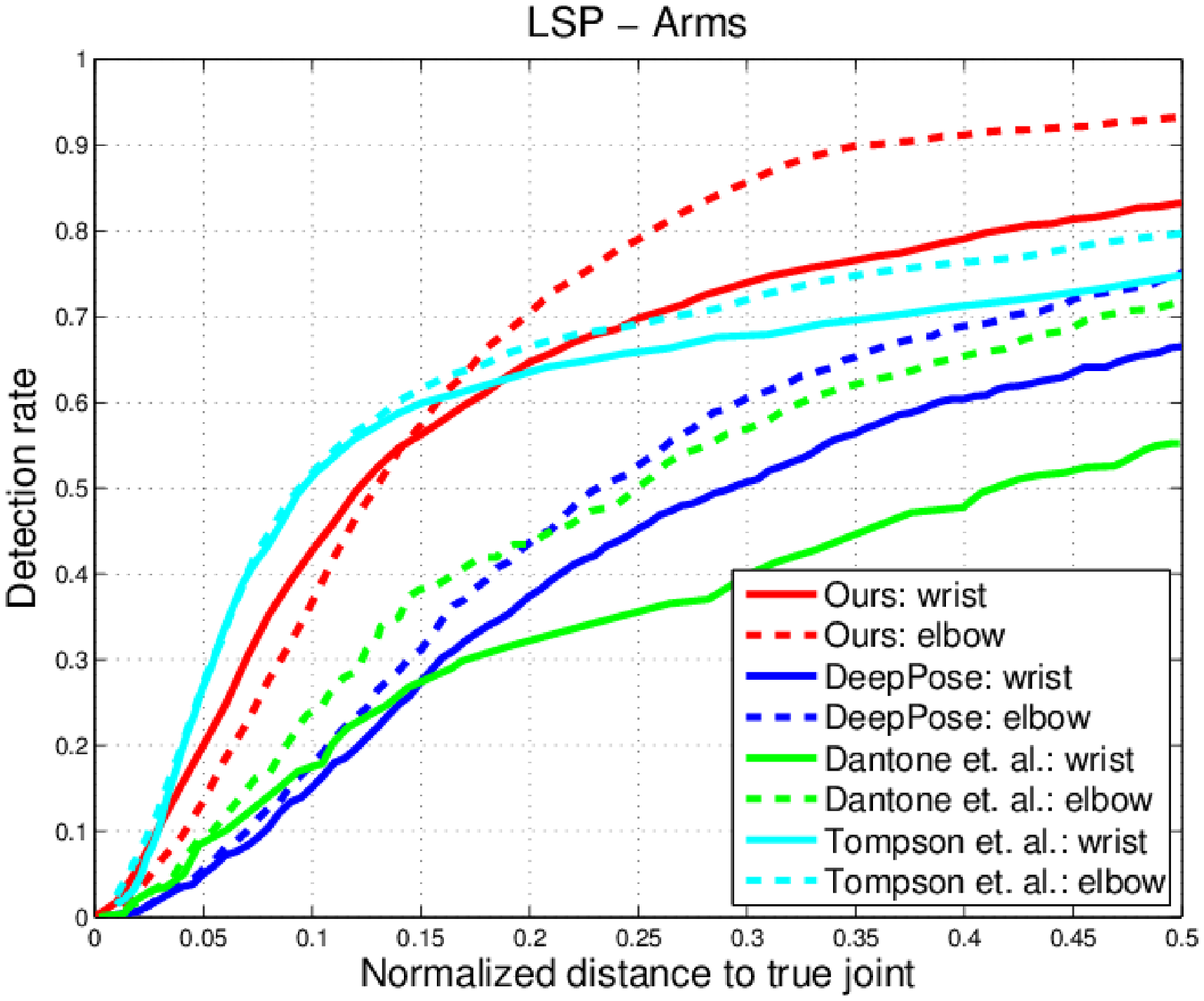}\includegraphics[width=0.23\textwidth]{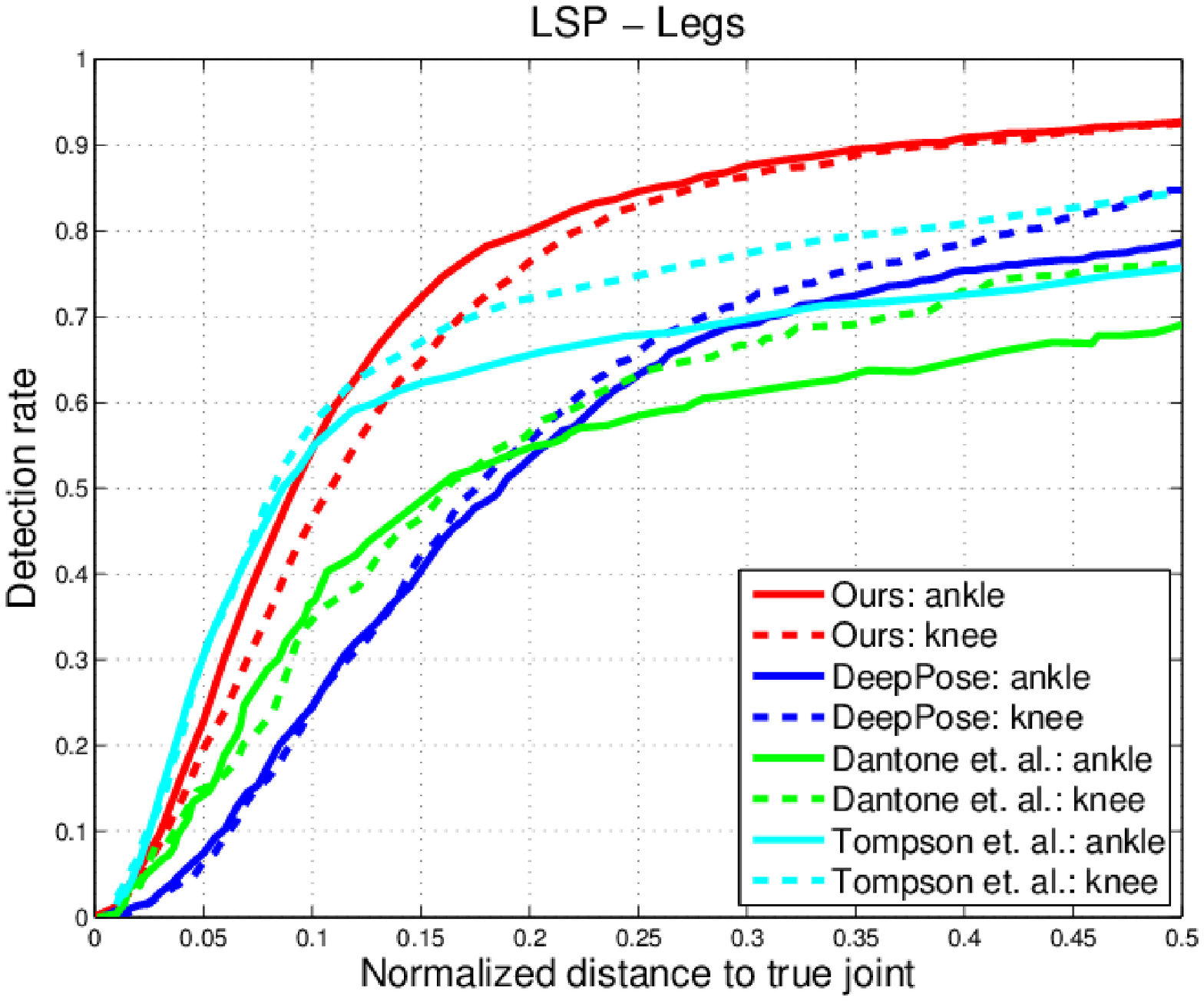}
\par\end{centering}

\begin{centering}
\caption{PDJ comparison on LSP. \label{fig:PDJs_on_LSP}}

\par\end{centering}

\centering{}
\vspace{-0.2in}
\end{figure}

Most LSP images only contain a single person. While each image in
FLIC may contain multiple people, similar to \cite{Toshev2014}, a
standard preprocessing of body detection has been conducted to extract
individual persons. As in previous works, we take the subimages of
these detected individual persons as training and testing samples.
This way, the training and testing data only contain \textit{a single}
\emph{person} and as mentioned earlier, in the testing stage, we simply
take the whole image (for FLIC dataset, this means a whole subimage
for an individual person) as the body patch. 

It has been verified that, in the training stage, the use of object
proposals can help train better CNNs for object detection and part
localization {[}17, 18{]}. However, in the testing stage, object proposals
detected on an image may be unevenly distributed. As a result, an
image region covered by dense low-likelihood object proposals may
undesirably show higher values in the resulting heatmap than a region
covered by sparser high-likelihood object proposals. To avoid this
issue, in our experiments we use multi-scale sliding windows (with
sizes of $0.5d\left(\mathbf{J}\right)$ and $d\left(\mathbf{J}\right)$,
stride 2) to provide part patches in the testing stage.

To compare with previous works, we evaluate the performance of human
pose estimation using two popular metrics: Percentage of Corrected
Parts (PCP) \cite{Eichner2012} and Percentage of Detected Joints
(PDJ) \cite{Sapp2013,Toshev2014}. PCP measures the rate of correct
limb detection -- a limb is detected if the distances between detected
limb joints and true limb joints are no more than half of the limb
length. Since PCP penalizes short limbs, PDJ is introduced to measure
the detection rate of joints, where a joint is considered to be detected
if the distance between detected joint and the true joint is less
than a fraction of the torso diameter $d\left(\mathbf{J}\right)$
as described in Section \ref{sec:Model-Inputs}. For PDJ, we can obtain
different detection rate by varying the fraction and generate a PDJ
curve in terms of the normalized distance to true joint \cite{Toshev2014}. 

The parameters that need to be set in the proposed method are 
\begin{enumerate}
\item Lower bound coefficient $\mu_{1}$ and the upper bound coefficient
$\mu_{2}$ in Eq.(\ref{eq:part_proposal_upper_bound}).
\item Balance factor $\lambda_{d}$ in the loss function in Eq. (\ref{eq:total_cost}).
\item $k$ and $\lambda_{h}$ that are used for selecting patch-pairs for
joint localization in Section \ref{sec:Human-Pose-Estimation}.
\end{enumerate}
In our experiments, we set $\mu_{1}=0.1$, $\mu_{2}=1.0$, $\lambda_{d}=4$,
$k=3$, and $\lambda_{h}=0.9$. 

In this paper, we use the open-source CNN library Caffe \cite{Jia2014}
for implementing DS-CNN. We finetune a CNN network pretrained on ImageNet
\cite{Krizhevsky2012} for training the proposed DS-CNN. Following
\cite{Girshick2014}, the learning rate is initialized to a tenth
of the initial ImageNet learning rate and is decreased by a factor
of ten after every certain number of iterations. 

\begin{figure}
\begin{centering}
\includegraphics[scale=0.12]{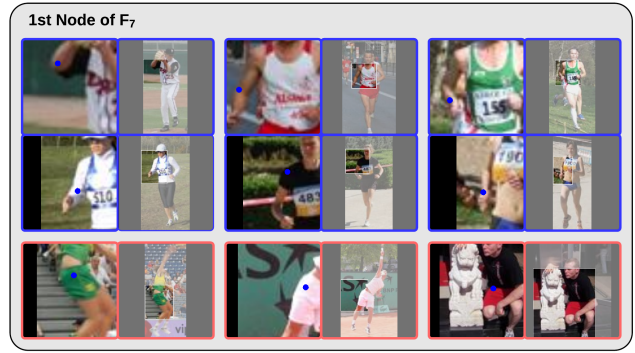}
\par\end{centering}

\caption{Visualization of the features extracted by layer ${\tt F}_{7}$ in
DS-CNN. \label{fig:Visualization-of-feat} }

\vspace{-0.0in}
\end{figure}

We first evaluate our method on LSP dataset. The PCP of the proposed
method, DeepPose and six other comparison methods for head, torso,
and four limbs (upper/lower arms and upper/lower legs) is shown in
Table \ref{tab:PCP-comparison}. Except for `head', our method outperforms
all the comparison methods including DeepPose at all body parts. The
improvement on average PCP is over 15\% against the best results obtained
by the comparison methods.

\begin{table*}
\begin{centering}

\par\end{centering}

\begin{centering}
\begin{tabular}{|c|>{\centering}p{0.7cm}|c|c|c|c|c|c|c|c||c|c|c|c|c|c|c|}
\hline 
{\footnotesize{}LSP} & {\scriptsize{}ankle} & {\scriptsize{}knee} & {\scriptsize{}hip} & {\scriptsize{}wrist} & {\scriptsize{}elbow} & {\scriptsize{}shoulder} & {\scriptsize{}neck} & {\scriptsize{}head } & {\scriptsize{}mAP} & {\scriptsize{}FLIC} & {\scriptsize{}hip} & {\scriptsize{}wrist} & {\scriptsize{}elbow} & {\scriptsize{}shoulder} & {\scriptsize{}Head } & {\scriptsize{}mAP}\tabularnewline
\hline 
\hline 
{\footnotesize{}$\mathbf{p}_{p}$} & {\footnotesize{}35.7} & {\footnotesize{}25.5} & {\footnotesize{}27.3} & {\footnotesize{}20.7} & {\footnotesize{}17.1} & {\footnotesize{}35.0} & {\footnotesize{}47.9} & {\footnotesize{}70.3} & {\footnotesize{}31.5} & {\footnotesize{}$\mathbf{p}_{p}$} & {\footnotesize{}61.2} & {\footnotesize{}56.0} & {\footnotesize{}71.2} & {\footnotesize{}88.8} & {\footnotesize{}93.8} & {\footnotesize{}72.0}\tabularnewline
\hline 
{\footnotesize{}$\mathbf{p}_{b}$} & {\footnotesize{}39.7} & {\footnotesize{}39.6} & {\footnotesize{}37.5} & {\footnotesize{}21.3} & {\footnotesize{}29.3} & {\footnotesize{}40.7} & {\footnotesize{}44.4} & {\footnotesize{}70.4} & {\footnotesize{}37.9} & {\footnotesize{}$\mathbf{p}_{b}$} & {\footnotesize{}72.8} & {\footnotesize{}59.3} & {\footnotesize{}77.7} & {\footnotesize{}91.0} & {\footnotesize{}94.0} & {\footnotesize{}77.2}\tabularnewline
\hline 
{\footnotesize{}$\mathbf{p}_{p,b}$} & \textbf{\footnotesize{}44.6} & \textbf{\footnotesize{}41.9} & \textbf{\footnotesize{}41.8} & \textbf{\footnotesize{}30.4} & \textbf{\footnotesize{}34.2} & \textbf{\footnotesize{}48.7} & \textbf{\footnotesize{}58.9} & \textbf{\footnotesize{}79.6} & \textbf{\footnotesize{}44.4} & {\footnotesize{}$\mathbf{p}_{p,b}$} & \textbf{\footnotesize{}74.3} & \textbf{\footnotesize{}68.1} & \textbf{\footnotesize{}82.0} & \textbf{\footnotesize{}93.5} & \textbf{\footnotesize{}96.4} & \textbf{\footnotesize{}81.4}\tabularnewline
\hline 
\end{tabular}
\par\end{centering}

\vspace{0.1in}

\caption{Average precision (\%) of joint detection on LSP and FLIC testing
datasets when CNN takes different types of patches as input. \label{tab:Detection-average-precision}}
\end{table*}

\begin{figure*}
\begin{centering}
\includegraphics[scale=1.2]{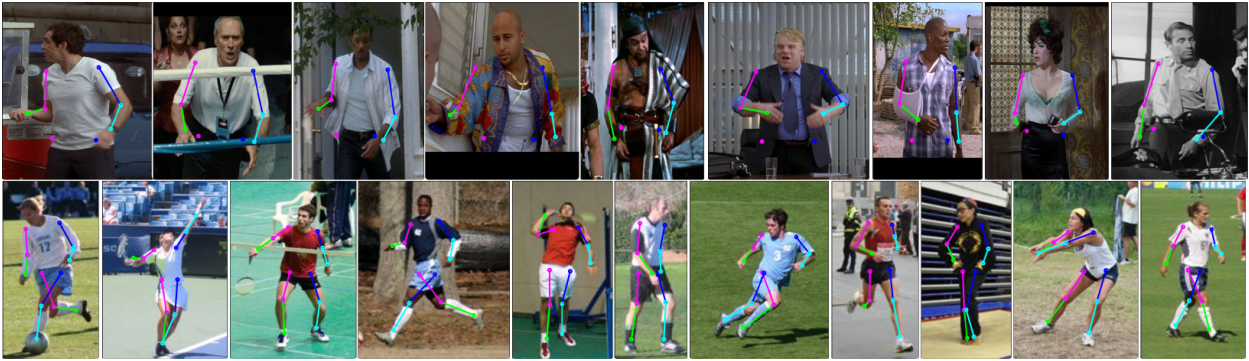}
\par\end{centering}

\caption{Human pose estimation on sample images from FLIC and LSP testing datasets.
\label{fig:Human-pose-estimation-examples}}

\vspace{-0.1in}
\end{figure*}

Figure \ref{fig:PDJs_on_FLIC} shows the PDJ curves of the proposed
method and seven comparison methods at the elbows and wrists on the
FLIC dataset \cite{Toshev2014,Sapp2013,Jain2014,Eichner2013,Sapp2010,Tompson2014,Yang2011}.
We can see that the proposed method outperforms all the comparison
methods except for Tompson et al. Tompson et al's PDJ performance
is higher than the proposed method, when normalized distance to true
joint, or for brevity, normalized distance, is less than a threshold
$t$, but a little lower than the proposed method when normalized
distance is larger than $t$. As shown in Fig. \ref{fig:PDJs_on_FLIC},
the value of $t$ is 0.15 and 0.18 for elbows and wrists respectively.

As a further note, Tompson et al. \cite{Tompson2014}  combines an
MRF-based graphic model into CNN-based part detection. It shows that
the inclusion of the graphical model can substantially improve the
PDJ performance. In this paper, we focus on developing a new CNN-based
method to detect local parts without using any high-level graphical
models. We believe the PDJ performance of the proposed method can
be further improved if we combine it to a graphical model as in Tompson
et al.

Performance comparison on LSP dataset using PDJ metric is shown in
Fig. \ref{fig:PDJs_on_LSP}. Similar to PDJ comparison on FLIC, the
PDJ of the proposed method is better than all the comparison methods
except for Tompson et al. When compared with Tompson et al, the proposed
method performs substantially better when normalized distance is large
and performs worse when normalized distance is small. One observation
is that the PDJ gain of the proposed method over Tompson et al. at
large normalized distance in LSP is more significant than the same
gain in FLIC. 

We also conduct an experiment to verify the effectiveness of using
dual sources of inputs: $\mathbf{p}_{p}$ and $\mathbf{p}_{b}$. In
this experiment, we compute the average precision (AP) of the joint
detection when taking either 1) only part patches $\mathbf{p}_{p}$
, 2) only body patches $\mathbf{p}_{b},$ or 3) the proposed patch
pairs $\mathbf{p}_{p,b}$ as the input to CNN. The results are shown
in Table \ref{tab:Detection-average-precision}. On both LSP and FLIC
testing datasets, the use of the dual-source patch-pairs achieves
better AP at all joints, and the best mAP, the average AP over all
the joints. Note that the body patch $\mathbf{p}_{b}$ in this paper
actually include part patch information, in the form of a binary mask
as discussed in Section \ref{sec:Model-Inputs}. That's why the use
of only $\mathbf{p}_{b}$ can lead significantly better AP than the
use of only $\mathbf{p}_{p}$  on both LSP and FLIC testing datasets.
However, the binary mask is usually of very low resolution because
we normalize the body patch to a fixed dimension. As a result, we
still need to combine $\mathbf{p}_{p}$ and $\mathbf{p}_{b}$ and
construct a dual-source CNN for pose estimation. 

Following \cite{Girshick2014,Ouyang2014}, we visualize the patterns
extracted by DS-CNN. We compute the activations in each hidden node
in layer ${\tt F}_{7}$ on a set of patch-pairs and Figure \ref{fig:Visualization-of-feat}
shows several patch pairs with the largest activations in the first
node of ${\tt F}_{7}$. We can see that this node fires for two pose
patterns -- the bent right elbow and the right hip. For each pattern,
the corresponding full-body pose also show high similarity because
of the inclusion of both part and body patches in DS-CNN.

Finally, sample human pose estimation results on FLIC and LSP testing
datasets are shown in Fig. \ref{fig:Human-pose-estimation-examples}.
In general, upper-body poses in FLIC are usually front-facing, while
the full-body pose in LSP contains many complex poses. As a result,
human pose estimation in LSP is less accurate than that in FLIC. By
including holistic views of part patches, the proposed method can
estimate the human pose even if some joints are occluded, as shown
in Fig. \ref{fig:Human-pose-estimation-examples}.

\section{Conclusion}

In this paper, we developed a new human pose estimation method based
on a dual-source convolutional neutral network (DS-CNN), which takes
two kinds of patches -- part patches and body patches -- as inputs
to combine both local and contextual information for more reliable
pose estimation. In addition, the output of DS-CNN is designed for
both joint detection and joint localization, which are combined for
estimating the human pose. By testing on the FLIC and LSP datasets,
we found that the proposed method can produce  superior performance
against several existing methods. When compared with Tompson et al
\cite{Tompson2014}, the proposed method performs better when normalized
distance is large and performs worse when normalized distance is small.
The proposed method is implemented using the open-source CNN library
Caffe and therefore has good expandability. \\
\\
\\

\textbf{Acknowledgement}: This work was supported in part by AFOSR FA9550-11-1-0327 and NSF IIS-1017199.

\begin{small}

\bibliographystyle{IEEEtran}
\bibliography{dual_source_cvpr15}

\end{small}
\end{document}